\documentclass[10pt,twocolumn,letterpaper]{article}
\usepackage{xcolor, soul}
\sethlcolor{orange}
\usepackage{xcolor}
\usepackage[normalem]{ulem}
\usepackage{parskip}
\usepackage{authblk}

\usepackage{cvpr}              

\usepackage{graphicx}
\usepackage{amsmath}
\usepackage{amssymb}
\usepackage{booktabs}
\usepackage{multirow}

\usepackage[pagebackref,breaklinks,colorlinks]{hyperref}

\usepackage[capitalize]{cleveref}
\crefname{section}{Sec.}{Secs.}
\Crefname{section}{Section}{Sections}
\Crefname{table}{Table}{Tables}
\crefname{table}{Tab.}{Tabs.}
\makeatletter \renewcommand\AB@affilsepx{\qquad \protect\Affilfont} 
\makeatother

\def\cvprPaperID{8} 
\def\confName{CVPR}
\def\confYear{2023}

\begin{document}

\title{Graph-CoVis: GNN-based Multi-view Panorama Global Pose Estimation }
\author[1]{Negar Nejatishahidin$^{*\dagger}$}
\author[2]{Will Hutchcroft$^*$}
\author[2]{Manjunath Narayana}
\author[2]{Ivaylo Boyadzhiev}
\author[2]{Yuguang Li}
\author[2]{Naji Khosravan}
\author[1]{Jana Košecká}
\author[2]{Sing Bing Kang}
\affil[1]{George Mason University}
\affil[2]{Zillow Group}

\maketitle
\def\thefootnote{*}\footnotetext{Equal contribution.}
\def\thefootnote{$\dagger$}\footnotetext{Done during Negar Nejatishahidin's internship at Zillow.}
\def\thefootnote{\arabic{footnote}}
\begin{abstract}
In this paper, we address the problem of wide-baseline camera pose estimation from a group of 360\protect{$^\circ$} panoramas under upright-camera assumption. Recent work has demonstrated the merit of deep-learning for end-to-end direct relative pose regression in 360\protect{$^\circ$} panorama pairs~\cite{Hutchcroft_2022}. To exploit the benefits of multi-view logic in a learning-based framework, we introduce Graph-CoVis, which non-trivially extends CoVisPose~\cite{Hutchcroft_2022} from relative two-view to global multi-view spherical camera pose estimation. Graph-CoVis is a novel Graph Neural Network based architecture that jointly learns the co-visible structure and global motion in an end-to-end and fully-supervised approach. Using the ZInD~\cite{Cruz2021ZillowID} dataset, which features real homes presenting wide-baselines, occlusion, and limited visual overlap, we show that our model performs competitively to state-of-the-art approaches.

\end{abstract}
\section{Introduction}
\label{sec:intro}

Camera pose estimation is a fundamental problem in computer vision and robotics. Whenever appropriate, constraints are used to both simplify the solution space and improve performance. One common constraint is that of planar camera motion. This is often the case when using sparsely captured $360^\circ$ panoramas for indoor applications. In our work, we address the multi-view pose estimation problem using $360^\circ$ panoramas with wide baselines within a large indoor space; we see this as a solution for an arbitrary number of visually connected panoramas.

\begin{figure}[t]
  \centering
    \includegraphics[width=1\linewidth]{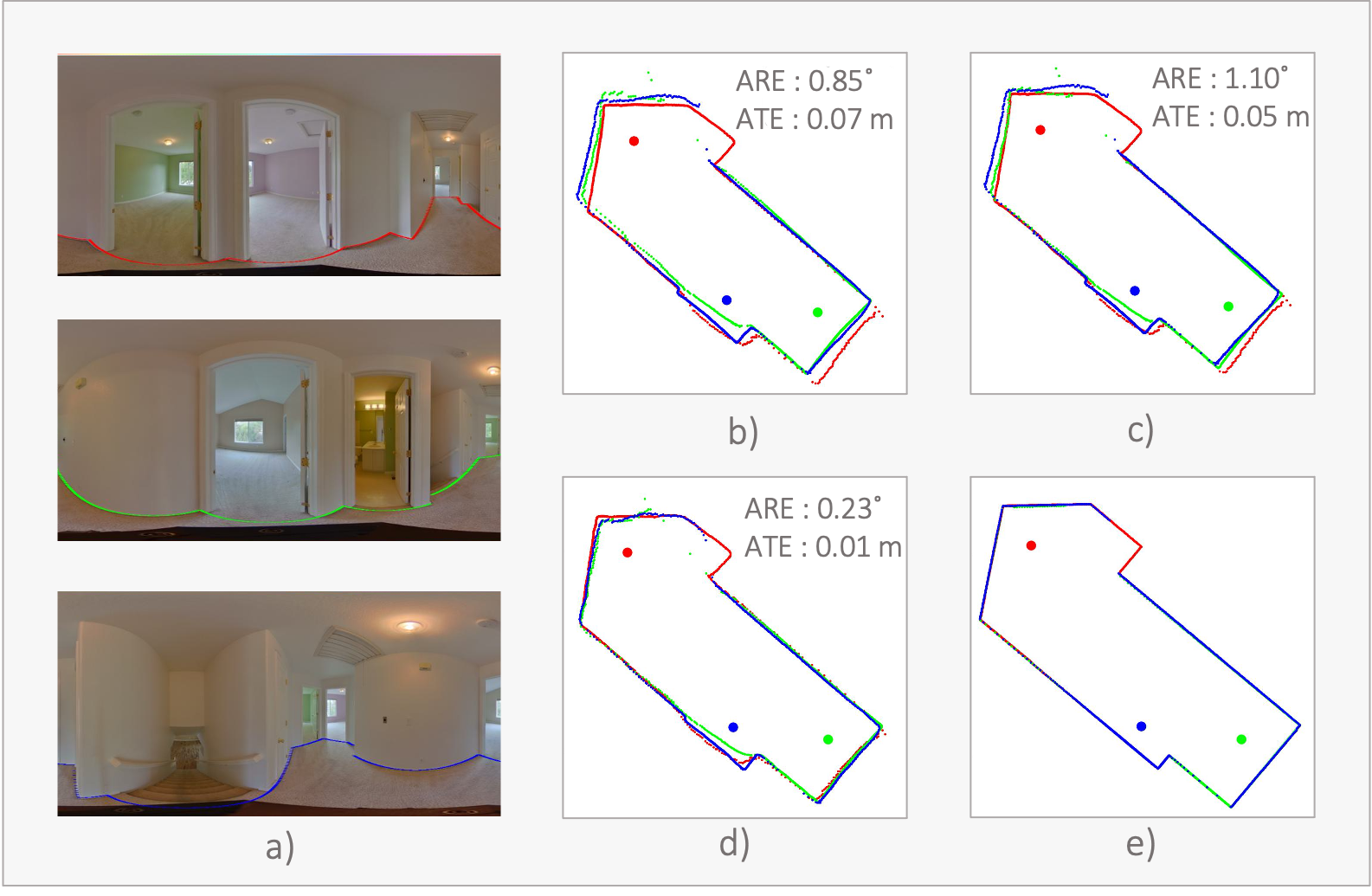}
    \caption{From three panoramas(a), Graph-CoVis(d) returns higher accuracy poses and visually consistent boundaries compared to two standard baselines based on relative pose estimation from CoVisPose\cite{Hutchcroft_2022} with greedy spanning tree(b) and pose graph optimization(c). Ground truth is shown in (e). ARE stands for absolute rotation error, and ATE stands for absolute translation error.}
   \label{fig:demo}
\end{figure}

CoVisPose~\cite{Hutchcroft_2022} is a state-of-the-art end-to-end model for pairwise relative pose estimation in $360^\circ$ indoor panoramas. It models the visual overlap and correspondence constraints that are present between two panoramic views when parts of an indoor scene are commonly observed by both cameras. In particular, by exploiting the upright-camera assumption, co-visibility (visual overlap), correspondence, and layout geometry are framed as 1-D quantities estimated over the image columns. Using this formulation, visualized in Figure~\ref{fig:demo}, CoVisPose demonstrates that learning such high-level geometric cues results in effective and robust representation for end-to-end pose estimation.




While CoVisPose achieves state-of-the-art results on wide-baseline relative pose estimation for pairs of 360\protect{$^\circ$} panoramas, it does not provide an end-to-end solution for more than two panoramas.
By comparison, we propose a more general end-to-end model for estimating the global poses for two or more panoramas. 

Our end-to-end approach, called Graph-CoVis, extends the strengths of the pair-wise pose estimation model from CoVisPose~\cite{Hutchcroft_2022} to a global multi-view pose estimation model. By using a Graph Neural Network (GNN)~\cite{Scarselli_2009}, Graph-CoVis retains the properties of the CoVisPose network that yield accurate pair-wise panorama pose estimates, while enabling it to generalize across multiple views and learn to regress consistent global poses.


Our technical contributions are as follow:
\begin{itemize}
    \item Graph-CoVis is the first end-to-end architecture for multi-view panorama global pose estimation.
    \item Graph-CoVis is the first global pose estimation architecture that can effectively handle varying numbers of input panoramas.
    \item A message passing scheme that enables the GNN to leverage deep representations of dense visual overlap and boundary correspondence constraints, to better estimate global pose.
    \item Competitive performance on ZInD for global pose estimation in a group of panoramas. 
\end{itemize}

\section{Related work}
\label{sec:related_work}
Estimation of the motion between two cameras is commonly achieved through detection and matching of keypoints such as SIFT\cite{Lowe2004DistinctiveIF} across the two images, estimating the transformation matrix between the two views, and finally computing the relative translation and rotation between the two cameras\cite{Hartley_2003}.
Commonly, RANSAC\cite{Fischler_1981} is used in the transformation estimation due to outliers in the matching stage.

\textbf{Learned models} have been proposed for each of the steps\cite{YiTLF16, Sarlin2020SuperGlueLF, Jiang2021COTRCT}, combination of steps\cite{ Dusmanu2019D2NetAT, Sun2021LoFTRDL}, and the end-to-end pipeline\cite{Parameshwara2022DiffPoseNetDD, Chen2021WideBaselineRC, Han2018RegNetLT, Musallam2022LeveragingEF}.
LIFT\cite{YiTLF16} is one of the first systems for a learned feature detector and descriptor, followed by more recent works such as SuperPoint\cite{ DeTone2018SuperPointSI},
Keypoint matching is learned using a GNN architecture in Superglue \cite{Sarlin2020SuperGlueLF} and a Transformer-based architecture in COTR\cite{Jiang2021COTRCT}. D2-Net\cite{Dusmanu2019D2NetAT} is a learned joint detector and descriptor.
Lofter\cite{Sun2021LoFTRDL} achieves detector-free matching across images by
learning feature descriptors starting from a dense pixel-wise sampling and refining them for high quality fine-level matching.
Differentiable RANSAC~\cite{BrachmannDSAC-CVPR2017} enables robustness in the end-to-end training of the pipeline.

\textbf{End-to-end methods} regress a pose directly from two input images. DiffPoseNet\cite{Parameshwara2022DiffPoseNetDD} learns poses by modeling optical flow and pose estimation, replicating these key principles from geometric methods.
Focusing on direction alone, DirectionNet\cite{Chen2021WideBaselineRC} works even for challenging wide-baseline images. RegNet\cite{Han2018RegNetLT} learns both the feature representations and the Jacobian matrix used in the optimization of two-view pose. Using a translation and rotation equivariant convolutional neural network\cite{Musallam2022LeveragingEF} improves the geometric information learned in the feature representations.
A common theme in these recent end-to-end approaches is the explicit modeling of two-view geometry principles in the network. 
Similarly, we leverage the strong geometry priors that are inherent in panorama images.


\textbf{GNNs} have been used for multi-view pose estimation in different ways. 
Similar in spirit to our work, PoGO-Net\cite{Li2021PoGONetPG} models multiple camera poses as nodes and uses a GNN with message passing scheme as an alternative to classical pose graph optimization. The method however requires an initialization method for the graph. In contrast, we do not require any explicit initialization. Our network densely connects each pose node to every other node and learns the dependencies between multiple views directly from the data.
\cite{Roessle2022End2EndMF} is an end-to-end trained GNN model to learn matches across multiple views, where the GNN module is followed by a differentiable pose optimization module. In contrast to our model that learns and updates the underlying features of the pair-wise module, their model is focused on learning the optimal matching function between keypoints.
Further, their model depends greatly on the accuracy of the pose optimizer to achieve good results. 

\begin{figure*}[t]
  \centering
   \includegraphics[width=\linewidth]{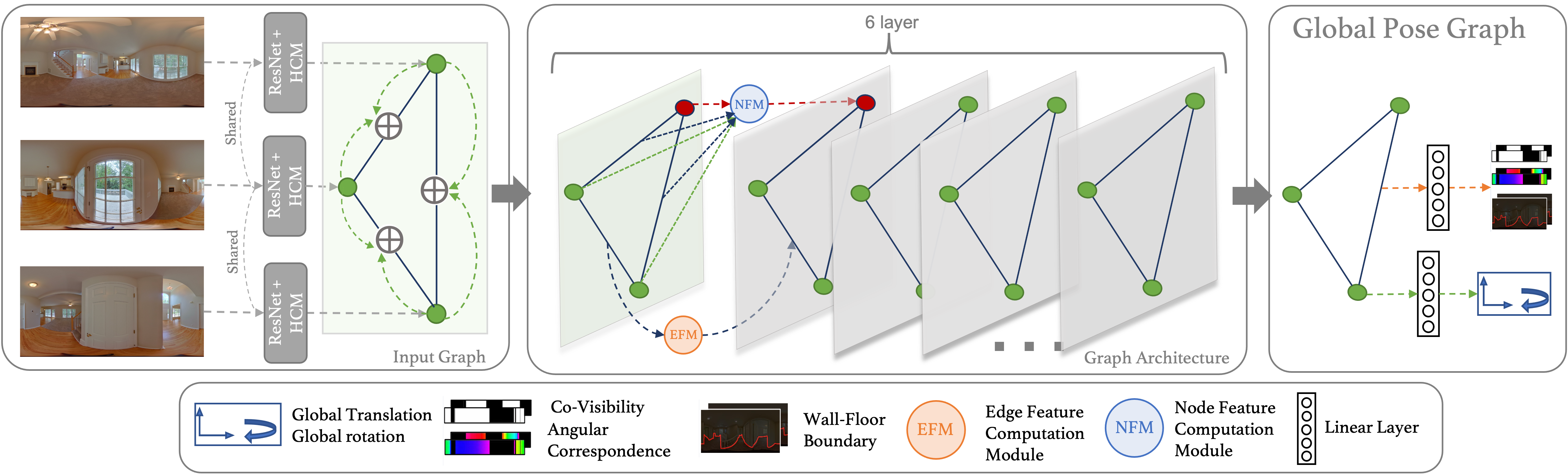}

   \caption{Graph-CoVis Architecture for a sample input of three panoramas. We initialize our graph's node and edge representations using visual features as in \cite{Hutchcroft_2022}, followed by six message passing layers to produce a Global Pose Graph. Nodes represent global poses and edges represent inter-frame geometric cues. The message passing process is: 1) the Edge Feature Computation Module (EFM) updates the edge features, 2) the Message Computation Module (MCM), where the target node's features attend to the features of the source node and the adjoining edge, and 3) the Node Feature Computation Module (NFM) aggregates the incoming messages from all source nodes to update target node features. Finally, each node and edge are used to estimate global poses and pair-wise geometric cues, respectively.}
   \label{fig:architecture}
\end{figure*}

Originally described for perspective images, some of the above methods have been applied on \textbf{panorama images}\cite{MurrugarraLlerena2022PoseEF}. 
The 360-degree view in panoramas creates useful constraints and angular correspondences between columns of two images that have a visual overlap between them. CoVisPose\cite{Hutchcroft_2022} leverages these constraints along with geometric priors in the appearance of structures such as room layout boundaries to yield a highly accurate two-view pose estimation model.
PSMNet\cite{Wang_2022_CVPR} is a pose and layout estimator that predicts the joint layout from two panorama views and is able to regress the fine pose when initialized to approximations. 

In our domain of \textbf{multi-view panorama} pose estimation notable recent works include estimating floor plans from extreme wide-baseline views (one panorama image per room)\cite{Shabani2021ExtremeSF} and SALVe\cite{Lambert2022SALVeSA}, a system for full floor plan reconstruction in sparsely sampled views. These works attempt to arrange all possible panoramas in the set, even those with little or no visual overlap, requiring alignment of semantic structures such as doors and walls. We address the problem of multi-view panorama pose estimation when panoramas have visual overlap between them, which is a key problem in full floor plan reconstruction.

\section{Method}
\label{sec:method}

In this section, we describe our Graph-CoVis architecture in detail, as well as our training strategy. The overall architecture for a group of three panoramas is visualized in Figure \ref{fig:architecture}. We start with briefly describing CoVisPose.

\subsection{CoVisPose}
CoVisPose\cite{Hutchcroft_2022} is an end-to-end method for pairwise pose estimation in wide baseline $360^\circ$ indoor panoramas. It uses geometric cues such as visible-boundary, co-visibility, and angular correspondence as auxiliary prediction outputs in order to effectively train a pose estimator. 
With features resulting from a ResNet backbone and a height compression module as input to the multi-layer transformer, it estimates the pose and geometric auxiliary outputs in separate branches. 

The Graph-CoVis framework generalizes the pair-wise pose estimation model to multiple views in order to estimate the global pose instead of the relative pose. In comparison to CoVisPose our model is capable of understanding global information inside the graph and using GNNs, we demonstrate the extension to multiple views by defining the following representations.
\subsection{Problem Representation}
Given a group of input panoramas of size $N$,  $\{\mathbf{I}_i\}_{i=1}^{N } \in \mathbb{R}^{3\times H \times W}$, without loss of generality we adopt $\mathbf{I}_{1}$ as the origin panorama, and estimate the remaining poses $\mathbf{P}_2$ to $\mathbf{P}_n$ in a shared coordinate system centered at the origin. 
We adopt the planar motion pose representation consisting of a translation vector $\mathbf{t} \in \mathbb{R}^2$ and a rotation matrix $R \in SO(2)$, i.e., the pose $\mathbf{P}_{i} \in SE(2)$. 
We represent the pose by 4 parameters, directly estimating the scaled translation vector $\mathbf{t}$, alongside the unit rotation vector $\mathbf{r}$.

\subsection{Graph Representation}






Defining the input directed graph as $\mathcal{G}=(\mathcal{V}, \mathcal{E})$, we represent the set of panoramas with nodes $\mathcal{V} = \{v_i\}$, and model the inter-image relationships through the edge set $\mathcal{E} = \left\{e_{ij} \mid v_i, v_j \in \mathcal{V} \right\}$. 

\subsubsection{Node and Edge Feature Initialization}
Each node $v_i$ in the graph $\mathcal G$ is associated with the node features $\mathbf{x}_{i}^{l}$, where $l$ refers to the layer number. The input graph node features, $\mathbf{x}_i^{0}$, are initialized with the visual features $\psi_i$, extracted from panorama $\mathbf{I}_i$. We employ the feature extractor from\cite{Hutchcroft_2022}, which consists of a ResNet50 backbone and a height compression module, followed by the addition of fixed positional encodings and six-layer transformer encoder. These structures are initialized from a pretrained CoVisPose model. 
Information about node identity is conveyed through learnable node embeddings.
The node embeddings also indicate to the network which node is the origin of the global coordinate frame.



The edge features $\mathbf{e}_{ij}^{0}$ are initialized with the concatenation of $\psi_i$ and $\psi_j$. Prior to concatenation, we additionally add the pretrained segment embeddings to identify the node identity for later edge feature computations. These convey image membership to the following transformer encoder layer. 


\subsection{Graph\-CoVis Network Architecture}




\subsubsection{Message Passing}
\begin{figure}[t]
  \centering
    \includegraphics[width=\linewidth]{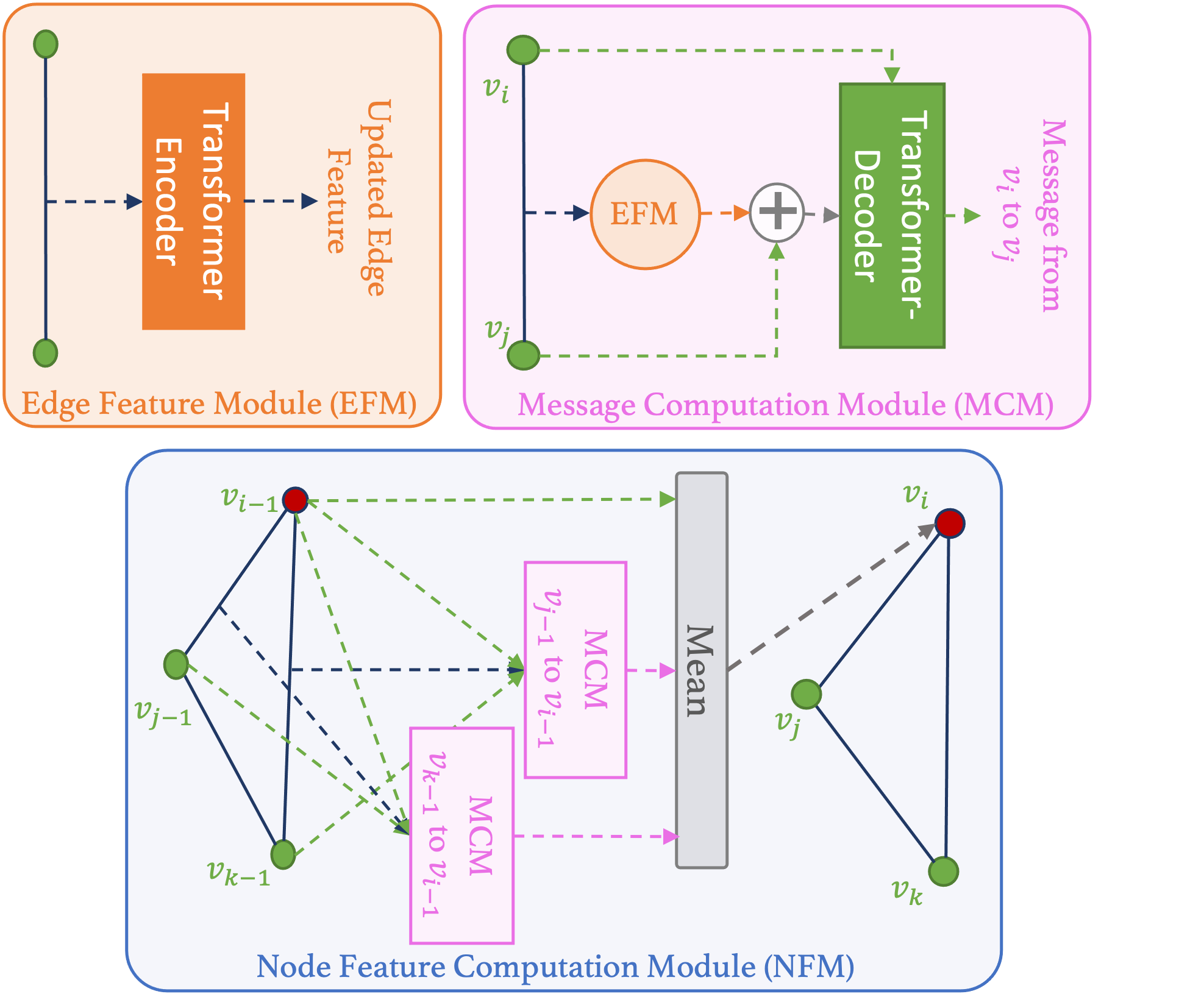}
    \caption{Graph-CoVis Message Passing. The Message Computation Module (MCM) computes incoming messages for each node. First, the Edge Feature Module (EFM) updates the edge representations with a single layer transformer. Then, the messages are computed through a transformer decoder, where the existing node representation attends to a concatenation of the edge representation and the adjacent nodes' embedding.}
   \label{fig:message_passing}
\end{figure}

Our network's representations are processed through six message passing layers to embed rich representations for pose regression. The message passing scheme is shown in Figure \ref{fig:message_passing}. We compute incoming messages for each node using the Message Computation Module (MCM).
The MCM first updates the edge features using the Edge Feature Module (EFM), and subsequently uses these representations to construct messages which are aggregated in the Node Feature Computation Module (NFM), to update the node embeddings. 

To update the edge features, the EFM 
consists of a single transformer encoder layer, the weights of which are initialized by the encoder layer weights from a pretrained CoVisPose model,
\begin{equation}
    \mathbf{e}_{ij}^{l} = \theta_E^l(\mathbf{e}_{ij}^{l-1}) ,
    \label{eq_edge_update}
\end{equation}
where $\theta_E^l$ is the single-layer transformer encoder in the $l$th message passing layer,  $\mathbf{e}_{ij}^{l-1}$ and $\mathbf{e}_{ij}^{l}$ are the edge features for edge $e_{ij}$ at the input and output of the EFM, respectively.
After the edge features have been updated in Eq. \ref{eq_edge_update}, the MCM then computes incoming messages for each node prior to aggregation using a single-layer transformer decoder, $\theta_M^l$,
\begin{equation}
    m_{j\rightarrow i}^l = \theta_M^l(\mathbf{x}_i^{l-1}, \mathbf{x}_j^{l-1}\oplus\mathbf{e}_{ij}^l) ,
    \label{eq_message_update}
\end{equation}
where $m_{j\rightarrow i}^l$ is the message from the source node $v_j$ to the target node $v_i$, and $\mathbf{x}_j^{l-1}\oplus\mathbf{e}_{ij}^l$ is the concatenation between the updated edge features $\mathbf{e}_{ij}^l$ and the existing node representation for the neighboring node $j$. In this way, the existing node representation attends to the inter-image information extracted along the edges, as well as the neighboring panoramas node representation. 

We subsequently update the node embeddings by taking the mean over all incoming messages in the Node Feature Computation Module (NFM), 
\begin{equation}
    \mathbf{x}_i^{l}=\dfrac{1}{deg(i)}\sum_{j \in \mathcal{N}(i)}m_{j\rightarrow i}^l ,
    \label{eq_node_update}
\end{equation}
where $j \in \mathcal{N}(i)$ represents the graph neighborhood of node $v_i$, and $deg(i)$ is the number of edges incident to node $v_i$.





\subsubsection{Co-Visibility, Angular Correspondence, and Floor-Wall Boundary}
We estimate dense column-wise representations of co-visibility, correspondence, and layout geometry similar to CoVisPose. Specifically, the edge features at the output of the final message passing layer are mapped to the dense column-wise outputs through a single fully connected layer, $\theta_{DC}$,
\begin{equation}
    [\boldsymbol{\phi}^{ij}, \boldsymbol{\alpha}^{ij}, \mathbf{p}^{ij}] = \theta_{DC}(\mathbf{e}_{ij}^L) ,
    \label{eq_dense}
\end{equation}
where $\boldsymbol{\phi}^{ij}$,$ \boldsymbol{\alpha}^{ij}, \mathbf{p}^{ij}$ are the column-wise vertical floor-wall boundary angle, angular correspondence, and co-visibility probability, respectively, and $\mathbf{e}_{ij}^L$ are the edge features at the output of the last layer, $L$. Again, we initialize $\theta_{DC}$ with weights from a pre-trained CoVisPose model. Learning these quantities along the edges encourages the edge features to embed information important for relative pose regression, which the node embeddings may then attend to in order to retain information relevant to global pose regression within the group of panoramas. 

\subsubsection{Pose Decoder}
To decode the node embeddings into the 4-parameter pose estimates, we apply three fully connected layers ($\theta_P$,), with Mish activation functions~\cite{Misra2020MishAS} between the first two layers. 
\begin{equation}
    [\mathbf{r}_i, \mathbf{t}_i] = \theta_P(\mathbf{x}_i^L).
    \label{eq_pose}
\end{equation}

\subsection{Training}
We train and evaluate our model on ZInD~\cite{Cruz2021ZillowID}, which is a large-scale dataset of real homes, containing multiple co-localized panoramas, with layout annotations necessary to support our layout-based correspondence and co-visibility representation. To create our dataset, we aggregate all spaces from ZInD containing three or more co-localized panoramas.
For the purpose of training and due to memory limitations, we set the maximum number of panoramas in a cluster to be five. 

During training, as large open spaces often contain much more than five panoramas, we randomly sample clusters of three, four, and five panoramas from these larger groups. For a set with $N$ panoramas, the model predicts $N$ global poses. Note that a single model is trained for all $N$ and the number of outputs from the model is determined by the number of input panoramas. We further apply random rotation augmentation, shifting the panoramas horizontally. Further, node ordering is permutated to yield a randomly selected origin node each time. 
Both types of augmentation result in altered coordinate systems and poses, presenting the network with varying pose targets during training. We use the publicly released train/test/validation split and train for 200 epochs, selecting the best model by validation error. 

\subsection{Loss Functions}
The loss function is composed of two main components, the node loss and the edge loss. The node loss, $\mathcal{L}_{\text{n}}$ itself consists of two terms  global node loss  $\mathcal{L}_{\text{ng}}$ and relative node loss $\mathcal{L}_{\text{nr}}$. We first directly minimize the pose error in a global coordinate system centered at the origin panorama through the global node loss,
\begin{equation}
    \mathcal{L}_{\text{ng}} = \sum_{i=2}^{N}(\|\mathbf{r}_i-\hat{\mathbf{r}_i}\|_2^2+\|\mathbf{t}_i-\hat{\mathbf{t}}_i\|_2^2) ,
    \label{eq_node_global_loss}
\end{equation}
where N represents the number of nodes in the graph. 

Additionally, the relative node loss is designed to encourage global consistency, we formulate this $\mathcal{L}_{\text{nr}}$ between all node estimates, $\hat{\mathbf{r}}_{ij}, \hat{\mathbf{t}}_{ij}$ and minimize the error against the ground truth relative poses.
This adds extra constraints between nodes other than the origin node. The relative pose node loss is
\begin{equation}
    \mathcal{L}_{\text{nr}} = \sum_{i}^{N}\sum_{j\neq i}^{N} (\|\mathbf{r}_{ij}-\hat{\mathbf{r}_{ij}}\|_2^2+\|\mathbf{t}_{ij}-\hat{\mathbf{t}}_{ij}\|_2^2) .
    \label{eq_node_relative_loss}
\end{equation}

The combined node loss is
\begin{equation}
    \mathcal{L}_{\text{n}} = \mathcal{L}_{\text{ng}} + \beta_r\cdot\mathcal{L}_{\text{nr}} ,
    \label{eq_node_loss}
\end{equation}
where $\beta_r$ is a constant controlling the relative influence of the global vs. relative pose losses, which we set to $0.1$.



The edge loss, $\mathcal{L}_{e}$, is applied to the dense co-visibility, correspondence, and layout geometry estimates as in\cite{Hutchcroft_2022},
\begin{equation}
    \mathcal{L}_{\text {e }}=\beta_{ac} \mathcal{L}_{ac}+\beta_{b} \mathcal{L}_b+\beta_{cv} \mathcal{L}_{cv}.
    \label{eq_edge_loss}
\end{equation}

The losses related to the other predicted outputs are,
\begin{equation}
    \mathcal{L}_{\text{b}} =
    \sum_{i=1}^N\sum_{j=1}^N\|\boldsymbol{\phi}_{ij}-\hat{\boldsymbol{\phi}}_{ij}\|_1, j \neq i
\end{equation}
\begin{equation}
    \mathcal{L}_{\text{ac}} =
    \sum_{i=1}^N\sum_{j=1}^N\|\boldsymbol{\alpha}_{ij}-\hat{\boldsymbol{\alpha}}_{ij}\|_1, j \neq i
\end{equation}
\begin{equation}
    \mathcal{L}_{\text{cv}} =
    \sum_{i=1}^N\sum_{j=1}^N BCE(\mathbf{p}_{ij}, \hat{\mathbf{p}}_{ij}), j \neq i ,
\end{equation}
where $\mathcal{L}_{\text{b}}, \mathcal{L}_{\text{ac}}, \mathcal{L}_{\text{cv}}$, are the layout boundary, angular correspondence, and co-visibility losses, respectively and $BCE$ is the binary cross entropy loss.


\subsection{Global origin selection}

During the training phase, the first panorama in the input list is considered the origin. At inference time, we run the model $\mathbf{N}$ times, with each panorama at the origin, retaining the result where the origin node has the highest mean co-visibility score to the neighboring panoramas.
\begin{table*}
\centering
    \begin{tabular}{@{}l*{7}{c}@{}}
        \toprule
        \multirow{2}{*}{Group-Size} & \multirow{2}{*}{Methods} & \multicolumn{3}{c}{Rotation} & \multicolumn{3}{c}{Translation} \\
        \cmidrule{3-8}
        && $\operatorname{Mn}\left({ }^{\circ} \downarrow\right)$ & $\operatorname{Med}\left({ }^{\circ} \downarrow\right) $ & $\operatorname{Std}({ }^{\circ} \downarrow)$ & $\operatorname{Mn}\left(\operatorname{m.} \downarrow\right)$ & $\operatorname{Med}\left(\operatorname{m.}  \downarrow\right)$ & $\operatorname{Std}(\operatorname{m.} \downarrow)$ \\
        \midrule
         
        \multirow{3}{*}{Three} & CoVisPose + Greedy& 2.648& 1.028&11.425 &0.093& 0.052&\textbf{0.244}\\
         & CoVisPose + PGO & 3.156& 0.984& 12.272&0.109& 0.047& 0.354\\
         & Graph-CoVis &\textbf{2.001}& \textbf{0.845}& \textbf{9.146} & \textbf{0.081}& \textbf{0.038}& 0.292\\\hline
        \multirow{2}{*}{Four} & CoVisPose + Greedy & 3.908& 1.161&16.557 & \textbf{0.142}& 0.068&\textbf{0.370}\\
        & CoVisPose + PGO &6.034& 1.310& 17.773&0.218& 0.067& 0.581\\
        & Graph-CoVis  & \textbf{3.192}& \textbf{0.941}& \textbf{13.359}& 0.153& \textbf{0.061}& 0.430\\\hline
        \multirow{2}{*}{Five} & CoVisPose + Greedy & 3.490& 1.257&13.928 & \textbf{0.154}& \textbf{0.078}&\textbf{0.344} \\
        & CoVisPose + PGO & 8.281& 1.715& 18.830&0.282& 0.089& 0.619\\
        & Graph-CoVis &\textbf{3.294}& \textbf{1.073}& \textbf{12.037}& 0.172& 0.082& 0.384\\

        \bottomrule
    \end{tabular}
    \caption{ Statistics of the rotation and translation error based on ARE and ATE metrics on group of three, four, and five panoramas for presented baselines and Graph-Covis. }
    \label{tab:final}
\end{table*}

\section{Experiments}

\begin{table*}
\centering 
\resizebox{\textwidth}{!}{
    \begin{tabular}{@{}l*{10}{c}@{}}
        \toprule
         \multirow{2}{*}{\small{Group-Size}} & \multirow{2}{*}{\small{Connection}}  & \multirow{2}{*}{\#Test} &\multirow{2}{*}{Methods} & \multicolumn{3}{c}{Rotation} & \multicolumn{3}{c}{Translation} \\
        \cmidrule{6-11}
        &&&& $\operatorname{Mn}\left({ }^{\circ} \downarrow\right)$ & $\operatorname{Med}\left({ }^{\circ} \downarrow\right) $ & $\operatorname{Std}({ }^{\circ} \downarrow)$ & $\operatorname{Mn}\left(\operatorname{m.} \downarrow\right)$ & $\operatorname{Med}\left(\operatorname{m.}  \downarrow\right)$ & $\operatorname{Std}(\operatorname{m.} \downarrow)$ \\
        \midrule
        \multirow{6}{*}{Three} 
        &\multirow{3}{*}{\small{Partially}}  &\multirow{3}{*}{52}&\multirow{3}{*}{108}   & CoVisPose + Greedy& 7.849& 1.991&18.743 &\textbf{0.308}& \textbf{0.095}&\textbf{0.641}\\
         &&&& CoVisPose + PGO &15.744 & 7.971& 21.691 & 0.685 & 0.283 & 1.218\\
         &&&& Graph-CoVis &\textbf{5.362}& \textbf{1.364}& \textbf{15.923} & 0.340& 0.124& 0.993\\ 
         \cmidrule{2-10}
         &\multirow{3}{*}{\small{Fully}}  &\multirow{3}{*}{1203}&\multirow{3}{*}{2886}   & CoVisPose + Greedy& 2.423& 1.007&10.944 & 0.084& 0.051&0.205\\

         &&&& CoVisPose + PGO & 2.612& 0.948& 11.386 &0.084& 0.046& 0.228 \\
         &&&& Graph-CoVis &\textbf{1.856}& \textbf{0.833}& \textbf{8.706}& \textbf{0.069}& \textbf{0.037}& \textbf{0.208}\\ \hline

        \multirow{6}{*}{Four} & \multirow{3}{*}{\small{Partially}}&\multirow{3}{*}{108}&\multirow{3}{*}{236}  &CoVisPose + Greedy &\textbf{6.776}& 1.671& \textbf{21.307} &\textbf{0.267}& \textbf{0.089}&\textbf{0.638}\\
        
        &&&& CoVisPose + PGO   &16.573& 6.070& 25.851&0.585& 0.215& 1.061\\  
        &&&& Graph-CoVis  &9.008& \textbf{1.403}& 25.240& 0.386& 0.137& 0.811\\ \cmidrule{2-10}
        
        & \multirow{3}{*}{\small{Fully}} &\multirow{3}{*}{437}&\multirow{3}{*}{1160} & CoVisPose + Greedy & 3.199& 1.069&15.071& 0.111& 0.064&0.256\\
        &&&& CoVisPose + PGO & 3.429& 1.045& 13.949 &0.127& 0.056& 0.319\\ 
        &&&&  Graph-CoVis &  \textbf{1.754}& \textbf{0.870}& \textbf{7.397}& \textbf{0.095}& \textbf{0.052}& \textbf{0.226}\\ \hline

        \multirow{6}{*}{Five} & \multirow{3}{*}{\small{Partially}}&\multirow{3}{*}{133} &\multirow{3}{*}{371}& CoVisPose + Greedy &4.996& 1.575&16.529 &\textbf{0.210}& \textbf{0.095}& \textbf{0.459}\\ 
        &&&& CoVisPose + PGO   &16.371& 6.528& 23.914 &0.518& 0.229& 0.838\\ 
        &&&& Graph-CoVis  &\textbf{4.713}& \textbf{1.320}& \textbf{13.568} & 0.246& 0.126& 0.462\\\cmidrule{2-10}
        
        & \multirow{3}{*}{\small{Fully}} &\multirow{3}{*}{219}&\multirow{3}{*}{609}& CoVisPose + Greedy&2.584& 1.107&11.986& \textbf{0.120}& 0.070&\textbf{0.244}\\
        &&&& CoVisPose + PGO   &3.368& 1.028& 12.599 &0.139& \textbf{0.063}& 0.367\\ 
        &&&& Graph-CoVis  & \textbf{2.433}& \textbf{0.948}& \textbf{10.915} &  0.128& 0.064& 0.319\\

        \bottomrule
    \end{tabular}}
    \caption{ Mean rotation and translation error for groups of three, four, and five panoramas divided into Fully and Partially co-visible sub-sets. The number of training and test examples are shown for each sub-set.}
    \label{tab:groups_345_partially_fully}
\end{table*}
\begin{figure*}[ht!]
  \centering
    \includegraphics[width=0.985\linewidth]
    {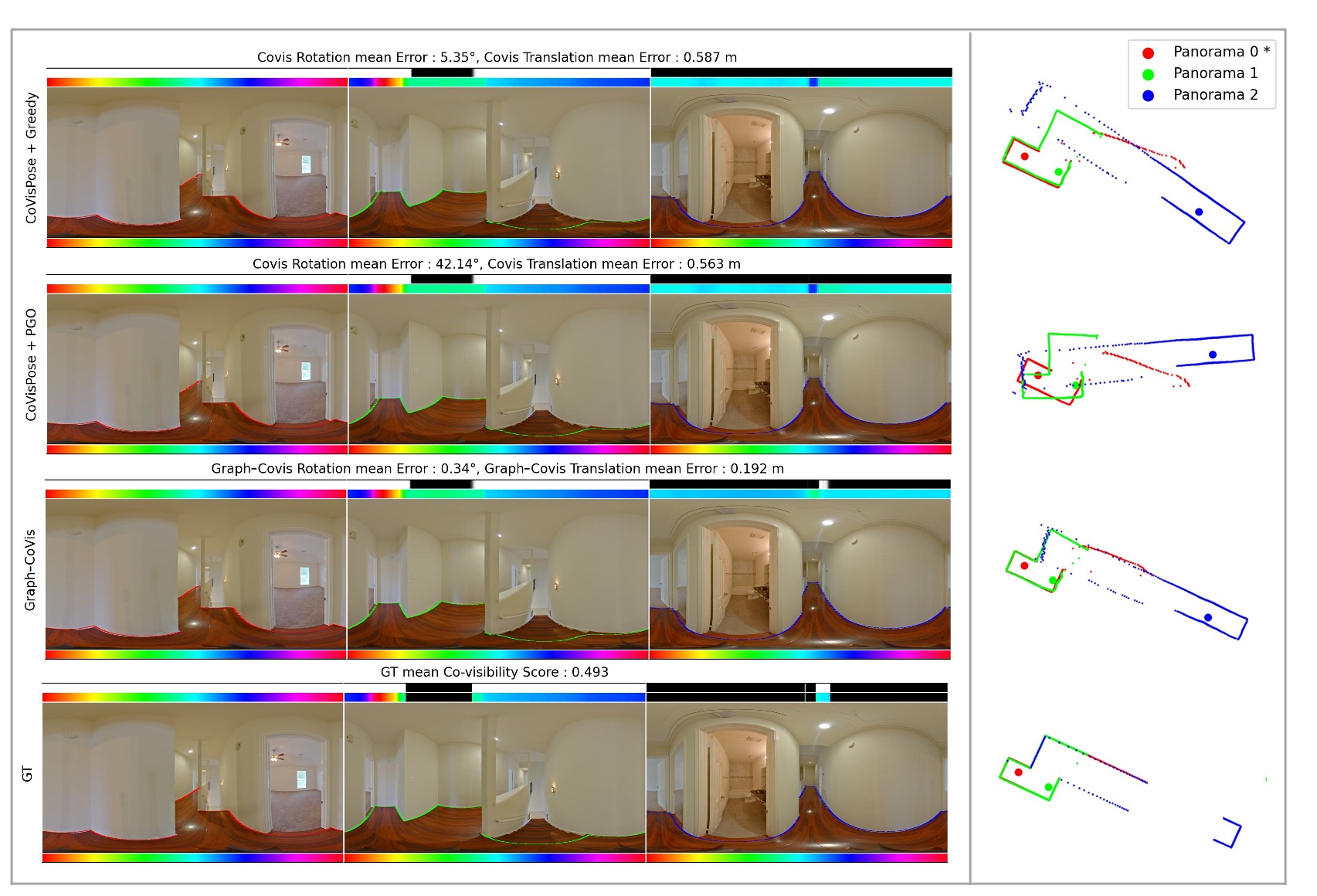}
     \caption{Visualization for both baselines and Graph-CoVis model. The red node represents the common origin node for all approaches. Graph-covis shows improvement in the mean rotation, translation error, and top-down alignment of predicted room boundaries. More examples are in the supplementary material.}
   \label{fig:Visualizations6}   
\end{figure*}
We compared our model against standard ways of extending the pair-wise pose estimates to multiple views with experiments on the ZInD data set.


\subsection{Baseline}
\label{Baseline}
We compare our model to two baseline methods based on the most recent and accurate pose estimation model for panorama images. CovisPose\cite{Hutchcroft_2022} has been demonstrated to be significantly better than alternatives for the domain of upright panorama images, under planar camera motion.

Taking a graph view of the problem with global poses representing nodes and relative pair-wise poses representing edges, we use two standard methods to extend pair-wise relative pose estimates from the CoVisPose model.

\textbf{Greedy spanning tree.} We sort the pair-wise poses by their predicted covisibility and add them greedily from highest covisibility to lowest until all panoramas are placed in the graph. This baseline is called CoVisPose + Greedy.

\textbf{Pose graph optimization.} The most common method to estimate global poses with multiple relative pair-wise poses is to use pose graph optimization (PGO)\cite{Dellaert12_GTSAM}. We use the graph structure from the greedy spanning tree baseline along with the edge that was not considered (lowest covisibility relative pose) as the pose graph and perform optimization. We call this baseline CoVisPose + PGO.


\subsection{Evaluation Metric}
To compute the error between ground truth and predicted poses for the panoramas, which are in arbitrary coordinate frames, we compute an alignment transformation between the two configurations. Using a least squares fit\cite{Arun_1987, Wahba_1965} to align the 2D point-sets ($x_i$ and $y_i$ locations of each panorama $i$ in the triplet), we first estimate a transformation matrix (rotation and translation in 2D space) to best align the ground truth and predicted poses. The difference between the positions and orientations of the aligned poses are reported as \emph{absolute translation error (ATE)} and  \emph{absolute rotation error (ARE)}.


\label{metric}

\subsection{Quantitative Results}
The results for mean, median, and standard deviation of the ARE and ATE,  separated by the number of panoramas in the set, are shown in Table \ref{tab:final}. Graph-CoVis performs better than the baselines for group size of three. For group size of four and five, Graph-CoVis performs better in rotation, but comparable or slightly worse in translation. 
While PGO moderately reduces the median translation error for group size three and four, PGO performs slightly worse than the greedy method with respect the other metrics. We hypothesize that this is because we ignore the least co-visibility relative pose in the greedy method. Low co-visibilty estimates are also more likely to be erroneous outliers. Including them affects PGO negatively.

To better understand the relation between the graph structure and model, we consider two cases of the connectivity between nodes.
Considering the visual overlap between nodes and removing connections between two nodes if overlap is less than a threshold of $0.1$, we have two possible graph structures: \emph{fully} and \emph{partially} connected sets.

Two panoramas are deemed to be connected if there is visual connectivity between them. A set of panoramas is fully connected if every pair in that set is visually connected, i.e., the ground truth co-visibility~\cite{Hutchcroft_2022} between them is greater than the threshold. It is partially connected if there is at least one pair that is not visually connected.
Table~\ref{tab:groups_345_partially_fully} shows that in general, Graph-Covis performs better than the baselines for \emph{Fully} connected examples. The table also shows an interesting correlation between accuracy and the number of training examples in each set.

\begin{figure*}
  \centering
    \includegraphics[width=1.03\linewidth]
    {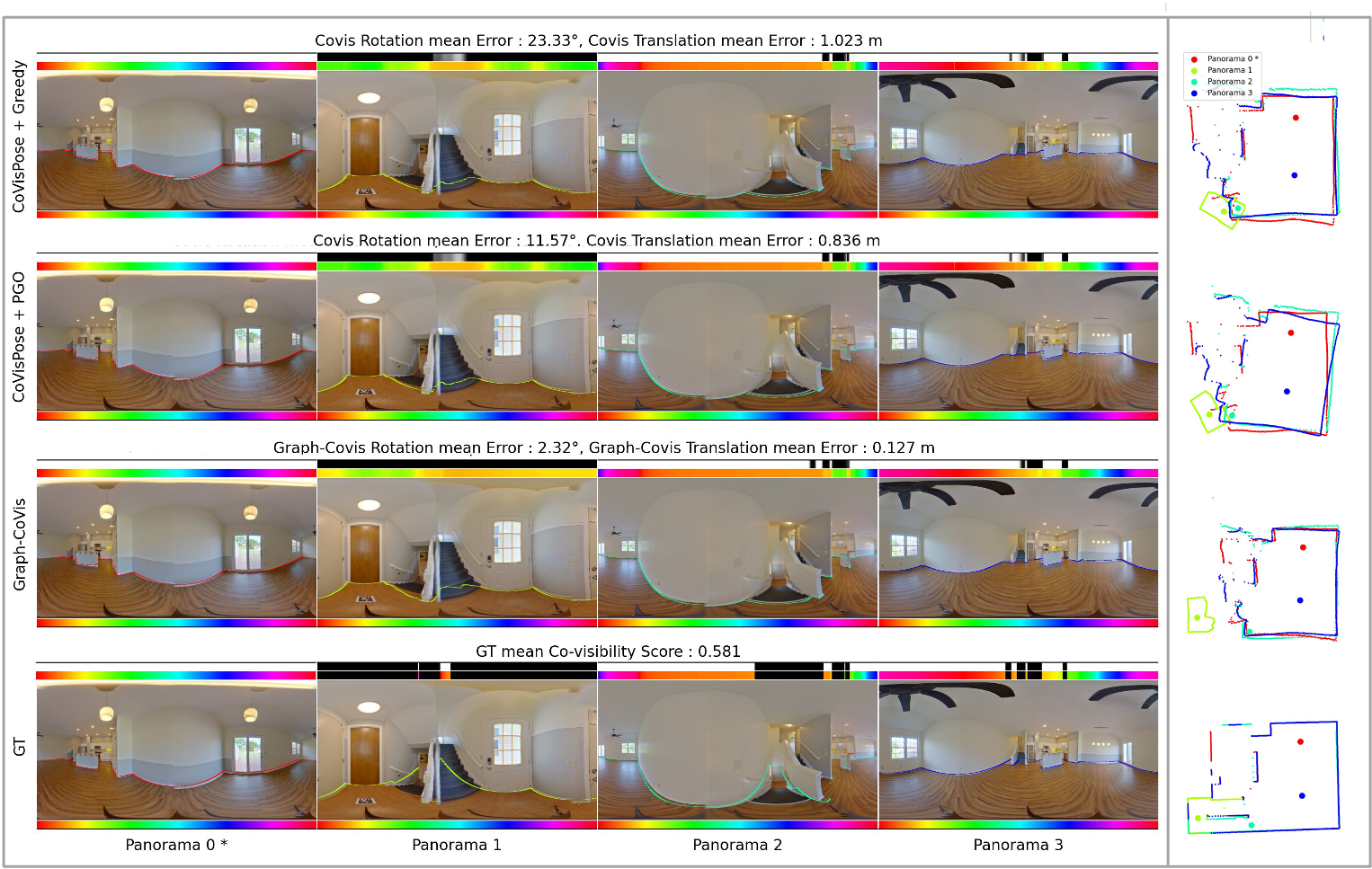}
    \caption{An example result for cluster size four.}
   \label{fig:Visualizations_four}
\end{figure*}

\begin{figure*}
  \centering
    \includegraphics[width=1.03\linewidth]
    {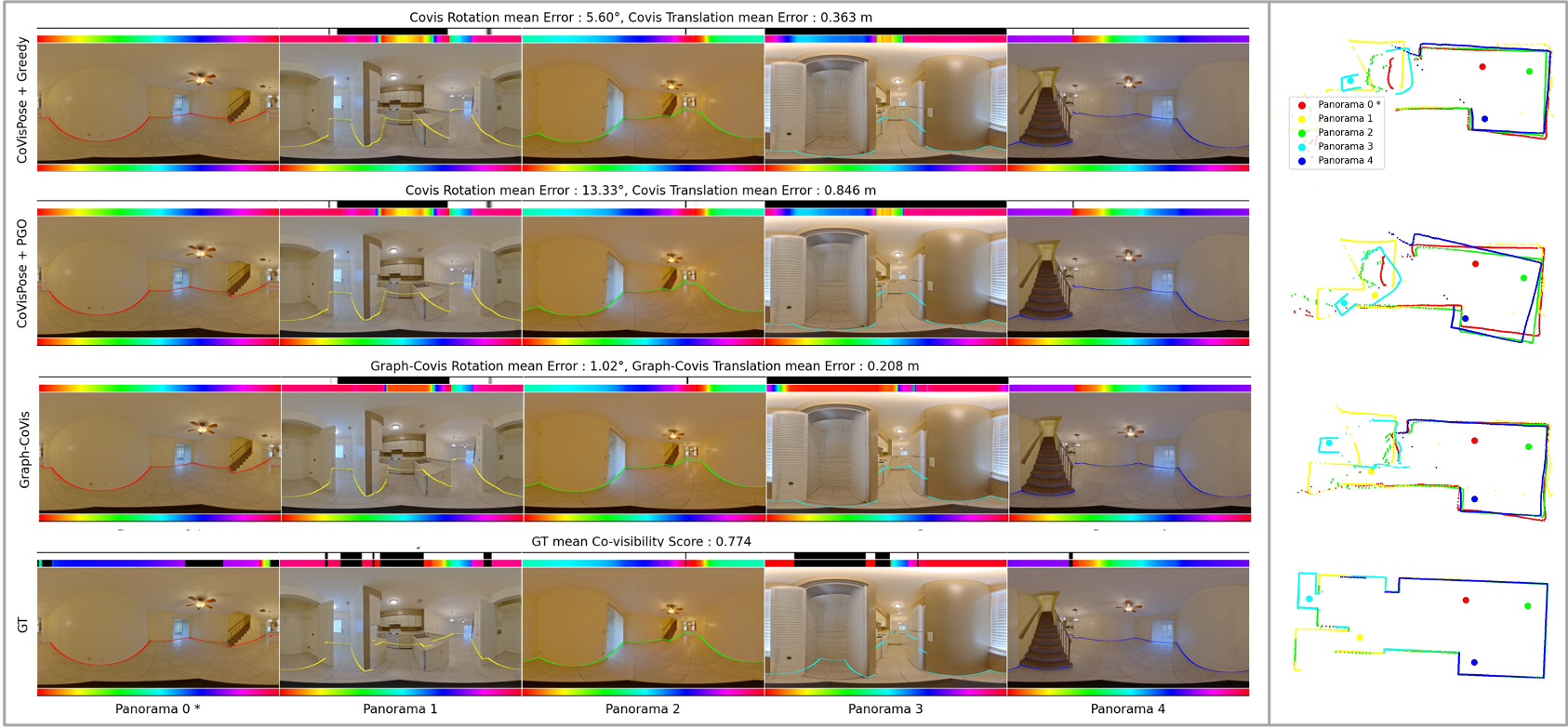}
    \caption{An example result for cluster size five.}
   \label{fig:Visualizations_five}
\end{figure*}
\subsection{Qualitative Results}
Figure \ref{fig:Visualizations6} shows a typical example triplet and the predicted pose and geometry from the baseline methods and Graph-CoVis. The first column is the panorama, selected as origin node. Above each image the binary strip indicates the predicted co-visibility to the the origin panorama. The color strips at the top (and bottom) of each image indicate the matching angular correspondence from the current panorama to the origin panorama (and origin panorama to current panorama).
Predicted boundaries are shown in colored lines within each image. The top-down view of the panorama poses (large dots) and the boundary predictions are shown in the last column. The rows correspond to CoVisPose + Greedy, CoVisPose + PGO, Graph-CoVis, and ground truth. Graph-Covis results in more accurate placement of the panorama poses as well as predicted boundary points. 

Figures \ref{fig:Visualizations_four} and \ref{fig:Visualizations_five} show inference examples of having four and five panoramas in the group.

\section{Limitations}
The CoVisPose representations of dense co-visibility, angular correspondence, and layout boundary, required to train our method, are derived from annotated room layouts, which are not available for some datasets. This limitation precludes the application of our method in absence of re-annotation. Further, these representations, as well as the planar motion model used by our method, exploit the upright camera and fixed camera height assumptions. As a result, our method is not directly applicable to handheld captures. 

\section{Conclusion}
We show that Graph-Covis is a generalization of two-view panorama pose estimation to multi-views. It results in an end-to-end trainable network that directly predicts global poses from an input set of images.  

\long\def\/*#1*/{}

\/*
\subsection{Language}

All manuscripts must be in English.

\subsection{Dual submission}

Please refer to the author guidelines on the \confName\ \confYear\ web page for a
discussion of the policy on dual submissions.

\subsection{Paper length}
Papers, excluding the references section, must be no longer than eight pages in length.
The references section will not be included in the page count, and there is no limit on the length of the references section.
For example, a paper of eight pages with two pages of references would have a total length of 10 pages.
{\bf There will be no extra page charges for \confName\ \confYear.}

Overlength papers will simply not be reviewed.
This includes papers where the margins and formatting are deemed to have been significantly altered from those laid down by this style guide.
Note that this \LaTeX\ guide already sets figure captions and references in a smaller font.
The reason such papers will not be reviewed is that there is no provision for supervised revisions of manuscripts.
The reviewing process cannot determine the suitability of the paper for presentation in eight pages if it is reviewed in eleven.

\subsection{The ruler}
The \LaTeX\ style defines a printed ruler which should be present in the version submitted for review.
The ruler is provided in order that reviewers may comment on particular lines in the paper without circumlocution.
If you are preparing a document using a non-\LaTeX\ document preparation system, please arrange for an equivalent ruler to appear on the final output pages.
The presence or absence of the ruler should not change the appearance of any other content on the page.
The camera-ready copy should not contain a ruler.
(\LaTeX\ users may use options of cvpr.sty to switch between different versions.)

Reviewers:
note that the ruler measurements do not align well with lines in the paper --- this turns out to be very difficult to do well when the paper contains many figures and equations, and, when done, looks ugly.
Just use fractional references (\eg, this line is $087.5$), although in most cases one would expect that the approximate location will be adequate.

\subsection{Paper ID}
Make sure that the Paper ID from the submission system is visible in the version submitted for review (replacing the ``*****'' you see in this document).
If you are using the \LaTeX\ template, \textbf{make sure to update paper ID in the appropriate place in the tex file}.

\subsection{Mathematics}

Please number all of your sections and displayed equations as in these examples:
\begin{equation}
  E = m\cdot c^2
  \label{eq:important}
\end{equation}
and
\begin{equation}
  v = a\cdot t.
  \label{eq:also-important}
\end{equation}
It is important for readers to be able to refer to any particular equation.
Just because you did not refer to it in the text does not mean some future reader might not need to refer to it.
It is cumbersome to have to use circumlocutions like ``the equation second from the top of page 3 column 1''.
(Note that the ruler will not be present in the final copy, so is not an alternative to equation numbers).
All authors will benefit from reading Mermin's description of how to write mathematics:
\url{http://www.pamitc.org/documents/mermin.pdf}.

\subsection{Blind review}

Many authors misunderstand the concept of anonymizing for blind review.
Blind review does not mean that one must remove citations to one's own work---in fact it is often impossible to review a paper unless the previous citations are known and available.

Blind review means that you do not use the words ``my'' or ``our'' when citing previous work.
That is all.
(But see below for tech reports.)

Saying ``this builds on the work of Lucy Smith [1]'' does not say that you are Lucy Smith;
it says that you are building on her work.
If you are Smith and Jones, do not say ``as we show in [7]'', say ``as Smith and Jones show in [7]'' and at the end of the paper, include reference 7 as you would any other cited work.

An example of a bad paper just asking to be rejected:
\begin{quote}
\begin{center}
    An analysis of the frobnicatable foo filter.
\end{center}

   In this paper we present a performance analysis of our previous paper [1], and show it to be inferior to all previously known methods.
   Why the previous paper was accepted without this analysis is beyond me.

   [1] Removed for blind review
\end{quote}

An example of an acceptable paper:
\begin{quote}
\begin{center}
     An analysis of the frobnicatable foo filter.
\end{center}

   In this paper we present a performance analysis of the  paper of Smith \etal [1], and show it to be inferior to all previously known methods.
   Why the previous paper was accepted without this analysis is beyond me.

   [1] Smith, L and Jones, C. ``The frobnicatable foo filter, a fundamental contribution to human knowledge''. Nature 381(12), 1-213.
\end{quote}

If you are making a submission to another conference at the same time, which covers similar or overlapping material, you may need to refer to that submission in order to explain the differences, just as you would if you had previously published related work.
In such cases, include the anonymized parallel submission~\cite{Authors14} as supplemental material and cite it as
\begin{quote}
[1] Authors. ``The frobnicatable foo filter'', F\&G 2014 Submission ID 324, Supplied as supplemental material {\tt fg324.pdf}.
\end{quote}

Finally, you may feel you need to tell the reader that more details can be found elsewhere, and refer them to a technical report.
For conference submissions, the paper must stand on its own, and not {\em require} the reviewer to go to a tech report for further details.
Thus, you may say in the body of the paper ``further details may be found in~\cite{Authors14b}''.
Then submit the tech report as supplemental material.
Again, you may not assume the reviewers will read this material.

Sometimes your paper is about a problem which you tested using a tool that is widely known to be restricted to a single institution.
For example, let's say it's 1969, you have solved a key problem on the Apollo lander, and you believe that the CVPR70 audience would like to hear about your
solution.
The work is a development of your celebrated 1968 paper entitled ``Zero-g frobnication: How being the only people in the world with access to the Apollo lander source code makes us a wow at parties'', by Zeus \etal.

You can handle this paper like any other.
Do not write ``We show how to improve our previous work [Anonymous, 1968].
This time we tested the algorithm on a lunar lander [name of lander removed for blind review]''.
That would be silly, and would immediately identify the authors.
Instead write the following:
\begin{quotation}
\noindent
   We describe a system for zero-g frobnication.
   This system is new because it handles the following cases:
   A, B.  Previous systems [Zeus et al. 1968] did not  handle case B properly.
   Ours handles it by including a foo term in the bar integral.

   ...

   The proposed system was integrated with the Apollo lunar lander, and went all the way to the moon, don't you know.
   It displayed the following behaviours, which show how well we solved cases A and B: ...
\end{quotation}
As you can see, the above text follows standard scientific convention, reads better than the first version, and does not explicitly name you as the authors.
A reviewer might think it likely that the new paper was written by Zeus \etal, but cannot make any decision based on that guess.
He or she would have to be sure that no other authors could have been contracted to solve problem B.
\medskip

\noindent
FAQ\medskip\\
{\bf Q:} Are acknowledgements OK?\\
{\bf A:} No.  Leave them for the final copy.\medskip\\
{\bf Q:} How do I cite my results reported in open challenges?
{\bf A:} To conform with the double-blind review policy, you can report results of other challenge participants together with your results in your paper.
For your results, however, you should not identify yourself and should not mention your participation in the challenge.
Instead present your results referring to the method proposed in your paper and draw conclusions based on the experimental comparison to other results.\medskip\\

\begin{figure}[t]
  \centering
  \fbox{\rule{0pt}{2in} \rule{0.9\linewidth}{0pt}}

   \caption{Example of caption.
   It is set in Roman so that mathematics (always set in Roman: $B \sin A = A \sin B$) may be included without an ugly clash.}
   \label{fig:onecol}
\end{figure}

\subsection{Miscellaneous}

\noindent
Compare the following:\\
\begin{tabular}{ll}
 \verb'$conf_a$' &  $conf_a$ \\
 \verb'$\mathit{conf}_a$' & $\mathit{conf}_a$
\end{tabular}\\
See The \TeX book, p165.

The space after \eg, meaning ``for example'', should not be a sentence-ending space.
So \eg is correct, {\em e.g.} is not.
The provided \verb'\eg' macro takes care of this.

When citing a multi-author paper, you may save space by using ``et alia'', shortened to ``\etal'' (not ``{\em et.\ al.}'' as ``{\em et}'' is a complete word).
If you use the \verb'\etal' macro provided, then you need not worry about double periods when used at the end of a sentence as in Alpher \etal.
However, use it only when there are three or more authors.
Thus, the following is correct:
   ``Frobnication has been trendy lately.
   It was introduced by Alpher~\cite{Alpher02}, and subsequently developed by
   Alpher and Fotheringham-Smythe~\cite{Alpher03}, and Alpher \etal~\cite{Alpher04}.''

This is incorrect: ``... subsequently developed by Alpher \etal~\cite{Alpher03} ...'' because reference~\cite{Alpher03} has just two authors.


\begin{figure*}
  \centering
  \begin{subfigure}{0.68\linewidth}
    \fbox{\rule{0pt}{2in} \rule{.9\linewidth}{0pt}}
    \caption{An example of a subfigure.}
    \label{fig:short-a}
  \end{subfigure}
  \hfill
  \begin{subfigure}{0.28\linewidth}
    \fbox{\rule{0pt}{2in} \rule{.9\linewidth}{0pt}}
    \caption{Another example of a subfigure.}
    \label{fig:short-b}
  \end{subfigure}
  \caption{Example of a short caption, which should be centered.}
  \label{fig:short}
\end{figure*}

\section{Formatting your paper}
\label{sec:formatting}

All text must be in a two-column format.
The total allowable size of the text area is $6\frac78$ inches (17.46 cm) wide by $8\frac78$ inches (22.54 cm) high.
Columns are to be $3\frac14$ inches (8.25 cm) wide, with a $\frac{5}{16}$ inch (0.8 cm) space between them.
The main title (on the first page) should begin 1 inch (2.54 cm) from the top edge of the page.
The second and following pages should begin 1 inch (2.54 cm) from the top edge.
On all pages, the bottom margin should be $1\frac{1}{8}$ inches (2.86 cm) from the bottom edge of the page for $8.5 \times 11$-inch paper;
for A4 paper, approximately $1\frac{5}{8}$ inches (4.13 cm) from the bottom edge of the
page.

\subsection{Margins and page numbering}

All printed material, including text, illustrations, and charts, must be kept
within a print area $6\frac{7}{8}$ inches (17.46 cm) wide by $8\frac{7}{8}$ inches (22.54 cm)
high.
Page numbers should be in the footer, centered and $\frac{3}{4}$ inches from the bottom of the page.
The review version should have page numbers, yet the final version submitted as camera ready should not show any page numbers.
The \LaTeX\ template takes care of this when used properly.

\subsection{Type style and fonts}

Wherever Times is specified, Times Roman may also be used.
If neither is available on your word processor, please use the font closest in
appearance to Times to which you have access.

MAIN TITLE.
Center the title $1\frac{3}{8}$ inches (3.49 cm) from the top edge of the first page.
The title should be in Times 14-point, boldface type.
Capitalize the first letter of nouns, pronouns, verbs, adjectives, and adverbs;
do not capitalize articles, coordinate conjunctions, or prepositions (unless the title begins with such a word).
Leave two blank lines after the title.

AUTHOR NAME(s) and AFFILIATION(s) are to be centered beneath the title
and printed in Times 12-point, non-boldface type.
This information is to be followed by two blank lines.

The ABSTRACT and MAIN TEXT are to be in a two-column format.

MAIN TEXT.
Type main text in 10-point Times, single-spaced.
Do NOT use double-spacing.
All paragraphs should be indented 1 pica (approx.~$\frac{1}{6}$ inch or 0.422 cm).
Make sure your text is fully justified---that is, flush left and flush right.
Please do not place any additional blank lines between paragraphs.

Figure and table captions should be 9-point Roman type as in \cref{fig:onecol,fig:short}.
Short captions should be centred.

\noindent Callouts should be 9-point Helvetica, non-boldface type.
Initially capitalize only the first word of section titles and first-, second-, and third-order headings.

FIRST-ORDER HEADINGS.
(For example, {\large \bf 1. Introduction}) should be Times 12-point boldface, initially capitalized, flush left, with one blank line before, and one blank line after.

SECOND-ORDER HEADINGS.
(For example, { \bf 1.1. Database elements}) should be Times 11-point boldface, initially capitalized, flush left, with one blank line before, and one after.
If you require a third-order heading (we discourage it), use 10-point Times, boldface, initially capitalized, flush left, preceded by one blank line, followed by a period and your text on the same line.

\subsection{Footnotes}

Please use footnotes\footnote{This is what a footnote looks like.
It often distracts the reader from the main flow of the argument.} sparingly.
Indeed, try to avoid footnotes altogether and include necessary peripheral observations in the text (within parentheses, if you prefer, as in this sentence).
If you wish to use a footnote, place it at the bottom of the column on the page on which it is referenced.
Use Times 8-point type, single-spaced.

\subsection{Cross-references}

For the benefit of author(s) and readers, please use the
{\small\begin{verbatim}
  \cref{...}
\end{verbatim}}  command for cross-referencing to figures, tables, equations, or sections.
This will automatically insert the appropriate label alongside the cross-reference as in this example:
\begin{quotation}
  To see how our method outperforms previous work, please see \cref{fig:onecol} and \cref{tab:example}.
  It is also possible to refer to multiple targets as once, \eg~to \cref{fig:onecol,fig:short-a}.
  You may also return to \cref{sec:formatting} or look at \cref{eq:also-important}.
\end{quotation}
If you do not wish to abbreviate the label, for example at the beginning of the sentence, you can use the
{\small\begin{verbatim}
  \Cref{...}
\end{verbatim}}
command. Here is an example:
\begin{quotation}
  \Cref{fig:onecol} is also quite important.
\end{quotation}

\subsection{References}

List and number all bibliographical references in 9-point Times, single-spaced, at the end of your paper.
When referenced in the text, enclose the citation number in square brackets, for
example~\cite{Authors14}.
Where appropriate, include page numbers and the name(s) of editors of referenced books.
When you cite multiple papers at once, please make sure that you cite them in numerical order like this \cite{Alpher02,Alpher03,Alpher05,Authors14b,Authors14}.
If you use the template as advised, this will be taken care of automatically.

\begin{table}
  \centering
  \begin{tabular}{@{}lc@{}}
    \toprule
    Method & Frobnability \\
    \midrule
    Theirs & Frumpy \\
    Yours & Frobbly \\
    Ours & Makes one's heart Frob\\
    \bottomrule
  \end{tabular}
  \caption{Results.   Ours is better.}
  \label{tab:example}
\end{table}

\subsection{Illustrations, graphs, and photographs}

All graphics should be centered.
In \LaTeX, avoid using the \texttt{center} environment for this purpose, as this adds potentially unwanted whitespace.
Instead use
{\small\begin{verbatim}
  \centering
\end{verbatim}}
at the beginning of your figure.
Please ensure that any point you wish to make is resolvable in a printed copy of the paper.
Resize fonts in figures to match the font in the body text, and choose line widths that render effectively in print.
Readers (and reviewers), even of an electronic copy, may choose to print your paper in order to read it.
You cannot insist that they do otherwise, and therefore must not assume that they can zoom in to see tiny details on a graphic.

When placing figures in \LaTeX, it's almost always best to use \verb+\includegraphics+, and to specify the figure width as a multiple of the line width as in the example below
{\small\begin{verbatim}
   \usepackage{graphicx} ...
   \includegraphics[width=0.8\linewidth]
                   {myfile.pdf}
\end{verbatim}
}

\subsection{Color}

Please refer to the author guidelines on the \confName\ \confYear\ web page for a discussion of the use of color in your document.

If you use color in your plots, please keep in mind that a significant subset of reviewers and readers may have a color vision deficiency; red-green blindness is the most frequent kind.
Hence avoid relying only on color as the discriminative feature in plots (such as red \vs green lines), but add a second discriminative feature to ease disambiguation.

\section{Final copy}

You must include your signed IEEE copyright release form when you submit your finished paper.
We MUST have this form before your paper can be published in the proceedings.

Please direct any questions to the production editor in charge of these proceedings at the IEEE Computer Society Press:
\url{https://www.computer.org/about/contact}.

*/
{\small
\bibliographystyle{ieee_fullname}
\bibliography{egbib}
}

\end{document}


\title{Graph-CoVis: GNN-based Multi-view Panorama Global Pose Estimation:\\Supplementary Material}

\author[1]{Negar Nejatishahidin$^{*\dagger}$}
\author[2]{Will Hutchcroft$^*$}
\author[2]{Manjunath Narayana}
\author[2]{Ivaylo Boyadzhiev}
\author[2]{Yuguang Li}
\author[2]{Naji Khosravan}
\author[1]{Jana Košecká}
\author[2]{Sing Bing Kang}
\affil[1]{George Mason University}
\affil[2]{Zillow Group}

\maketitle

\def\thefootnote{$*$}\footnotetext{Equal contribution.}\def\thefootnote{\arabic{footnote}}
\def\thefootnote{$\dagger$}\footnotetext{Done during Negar Nejatishahidin's internship at Zillow.}\def\thefootnote{\arabic{footnote}}
\section{Group size and connectivity}\label{sec:generalize_graph_covis}

To further investigate the results, we evaluate the performance of the model on the data set separated into group size (number of panoramas) and percentage of co-visible connections between the panoramas. Table~\ref{tab:groups_345_connect_percent} shows results of this analysis. 
The main takeaway is that the rotation and translation errors are low when there are a large number of training examples. 
For datasets with more than $500$ training examples, the maximum mean translation error for Graph-CoVis is $0.13m$, whereas the best performing baseline has a maximum error of $0.11m$. For datasets with fewer than $100$ training examples, the minimum mean translation error for Graph-CoVis is $0.341m$, whereas the best performing baseline has a minimum error of only $0.14m$. Thus, Graph-CoVis performs worse for datasets with very few training examples.

\begin{table*}
\resizebox{\textwidth}{!}{
    \begin{tabular}{@{}l*{10}{c}@{}}
        \toprule
        \multirow{2}{*}{Group-Size} & \multirow{2}{*}{\%\small{Connection}}  & \multirow{2}{*}{\#Test}& \multirow{2}{*}{\#Train} &\multirow{2}{*}{Methods} & \multicolumn{3}{c}{Rotation} & \multicolumn{3}{c}{Translation} \\
        \cmidrule{6-11}
        &&&&& $\operatorname{Mn}\left({ }^{\circ} \downarrow\right)$ & $\operatorname{Med}\left({ }^{\circ} \downarrow\right) $ & $\operatorname{Std}({ }^{\circ} \downarrow)$ & $\operatorname{Mn}\left(\operatorname{m.} \downarrow\right)$ & $\operatorname{Med}\left(\operatorname{m.}  \downarrow\right)$ & $\operatorname{Std}(\operatorname{m.} \downarrow)$ \\
        \midrule
        \multirow{6}{*}{Three} 
        &\multirow{3}{*}{66\%}  &\multirow{3}{*}{52} &\multirow{3}{*}{108} & CoVisPose + Greedy& 7.849& 1.991&18.743 &\textbf{0.308}& \textbf{0.095}&\textbf{0.641}\\
         &&&& CoVisPose + PGO &15.744 & 7.971& 21.691 & 0.685 & 0.283 & 1.218\\
         &&&& Graph-CoVis &\textbf{5.362}& \textbf{1.364}& \textbf{15.923} & 0.340& 0.124& 0.993\\ 
         \cmidrule{2-11}
         &\multirow{3}{*}{100\%}  &\multirow{3}{*}{1203} &\multirow{3}{*}{2886} & CoVisPose + Greedy& 2.423& 1.007&10.944 & 0.084& 0.051&0.205\\

         &&&& CoVisPose + PGO & 2.612& 0.948& 11.386 &0.084& 0.046& 0.228 \\
         &&&& Graph-CoVis &\textbf{1.856}& \textbf{0.833}& \textbf{8.706}& \textbf{0.069}& \textbf{0.037}& \textbf{0.208}\\ \hline
         

        \multirow{12}{*}{Four} & \multirow{3}{*}{50\%}&\multirow{3}{*}{3} &\multirow{3}{*}{10} &CoVisPose + Greedy &53.348& 34.067&60.672& 1.908& 1.642&\textbf{1.856}\\
        &&&& CoVisPose + PGO & 51.777& \textbf{25.058}& 59.321 & \textbf{1.614}& \textbf{0.867} & 1.936\\
        &&&& Graph-CoVis  &\textbf{45.675}& 27.136& \textbf{57.675}&2.638& 2.113& 2.478\\
        \cmidrule{2-11}
        & \multirow{3}{*}{66\%} &\multirow{3}{*}{38} &\multirow{3}{*}{73}& CoVisPose + Greedy  & \textbf{7.048}& \textbf{1.920}&\textbf{22.248}&\textbf{0.248}& \textbf{0.099}&0.598\\
        &&&& CoVisPose + PGO &21.626& 8.032& 28.184&  0.671& 0.277& 1.156 \\
        &&&&  Graph-CoVis  & 10.694& 2.099& 25.795&  0.391& 0.166& \textbf{0.567}\\ 
        \cmidrule{2-11}
        & \multirow{3}{*}{83\%} &\multirow{3}{*}{67} &\multirow{3}{*}{153}& CoVisPose + Greedy& \textbf{4.536}& 1.413&\textbf{13.568} &  \textbf{0.204}& \textbf{0.081}&\textbf{0.418}\\
        &&&& CoVisPose + PGO & 12.131& 4.962& 19.478& 0.490& 0.180& 0.910\\
        &&&&  Graph-CoVis  &6.411& \textbf{1.204}& 20.784& 0.282& 0.110& 0.603\\ 
        \cmidrule{2-11}
        & \multirow{3}{*}{100\%} &\multirow{3}{*}{437} &\multirow{3}{*}{1160}& CoVisPose + Greedy& 3.199& 1.069&15.071& 0.111& 0.064&0.256\\
        &&&& CoVisPose + PGO & 3.429& 1.045& 13.949 &0.127& 0.056& 0.319\\ 
        &&&&  Graph-CoVis &  \textbf{1.754}& \textbf{0.870}& \textbf{7.397}& \textbf{0.095}& \textbf{0.052}& \textbf{0.226}\\ \hline

        \multirow{18}{*}{Five} & \multirow{3}{*}{50\%} &\multirow{3}{*}{3}&\multirow{3}{*}{2} & CoVisPose + Greedy &\textbf{2.739}& \textbf{2.472}&\textbf{2.053}&\textbf{0.140}& \textbf{0.115}&\textbf{0.083}\\ 
        &&&& CoVisPose + PGO & 26.961& 5.718& 35.364&0.830&0.559& 0.777\\
        &&&&  Graph-CoVis  &8.262& 1.046& 10.001 &0.504& 0.418& 0.412\\
        \cmidrule{2-11}
        & \multirow{3}{*}{60\%} &\multirow{3}{*}{14} &\multirow{3}{*}{39}& CoVisPose + Greedy&10.203& \textbf{2.961}&18.762& 0.539&\textbf{ 0.145}&0.994\\ 
        &&&& CoVisPose + PGO &27.103& 12.960& 34.467& 0.810& 0.478& 0.956\\
        &&&&   Graph-CoVis & \textbf{9.439}& 3.542& \textbf{18.216} & \textbf{0.528}& 0.217& \textbf{0.924}\\
        \cmidrule{2-11}
        & \multirow{3}{*}{70\%} &\multirow{3}{*}{26} &\multirow{3}{*}{64}& CoVisPose + Greedy&\textbf{8.096}& 2.114&26.835&\textbf{0.196}& \textbf{0.132}&\textbf{0.212}\\ 
        &&&& CoVisPose + PGO & 20.244& 11.380&23.232&0.721&0.267& 1.190\\
        &&&&  Graph-CoVis & 8.303& \textbf{1.889}& \textbf{17.971} &  0.341& 0.162& 0.627\\
        \cmidrule{2-11}
        & \multirow{3}{*}{80\%} &\multirow{3}{*}{44} &\multirow{3}{*}{117}& CoVisPose + Greedy&  4.155& \textbf{1.211}&14.433& \textbf{0.169}& \textbf{0.080}&0.285\\ 
        &&&& CoVisPose + PGO & 18.725& 8.783& 24.781& 0.471&0.232& 0.658\\ 
        &&& &  Graph-CoVis & \textbf{3.311}& 1.236& \textbf{13.439} &  0.181& 0.106& \textbf{0.218}\\
        
        \cmidrule{2-11}
        & \multirow{3}{*}{90\%} &\multirow{3}{*}{46} &\multirow{3}{*}{149}& CoVisPose + Greedy &2.611& 1.516&7.474& 0.160& \textbf{0.078}&0.408\\ 
        &&&& CoVisPose + PGO & 7.975& 3.258& 13.829 &0.339& 0.138& 0.639\\
        &&& &  Graph-CoVis & \textbf{2.354}& \textbf{1.022}& \textbf{6.865}&\textbf{ 0.151}& 0.098& \textbf{0.181}\\
        \cmidrule{2-11}
        & \multirow{3}{*}{100\%} &\multirow{3}{*}{219} &\multirow{3}{*}{609}&CoVisPose + Greedy&2.584& 1.107&11.986& \textbf{0.120}& 0.070&\textbf{0.244}\\
        &&&& CoVisPose + PGO   &3.368& 1.028& 12.599 &0.139& \textbf{0.063}& 0.367\\ 
        &&&& Graph-CoVis  & \textbf{2.433}& \textbf{0.948}& \textbf{10.915} &  0.128& 0.064& 0.319\\

        \bottomrule

    \end{tabular}
    }
    \caption{ Mean rotation and translation error for different group sizes, separated into sub-sets based on connectivity (percentage) between panoramas. The number of training and test examples are shown for each sub-set.}
    \label{tab:groups_345_connect_percent}
\end{table*}

\begin{figure*}[t]
  \centering
    \includegraphics[width=1\linewidth]{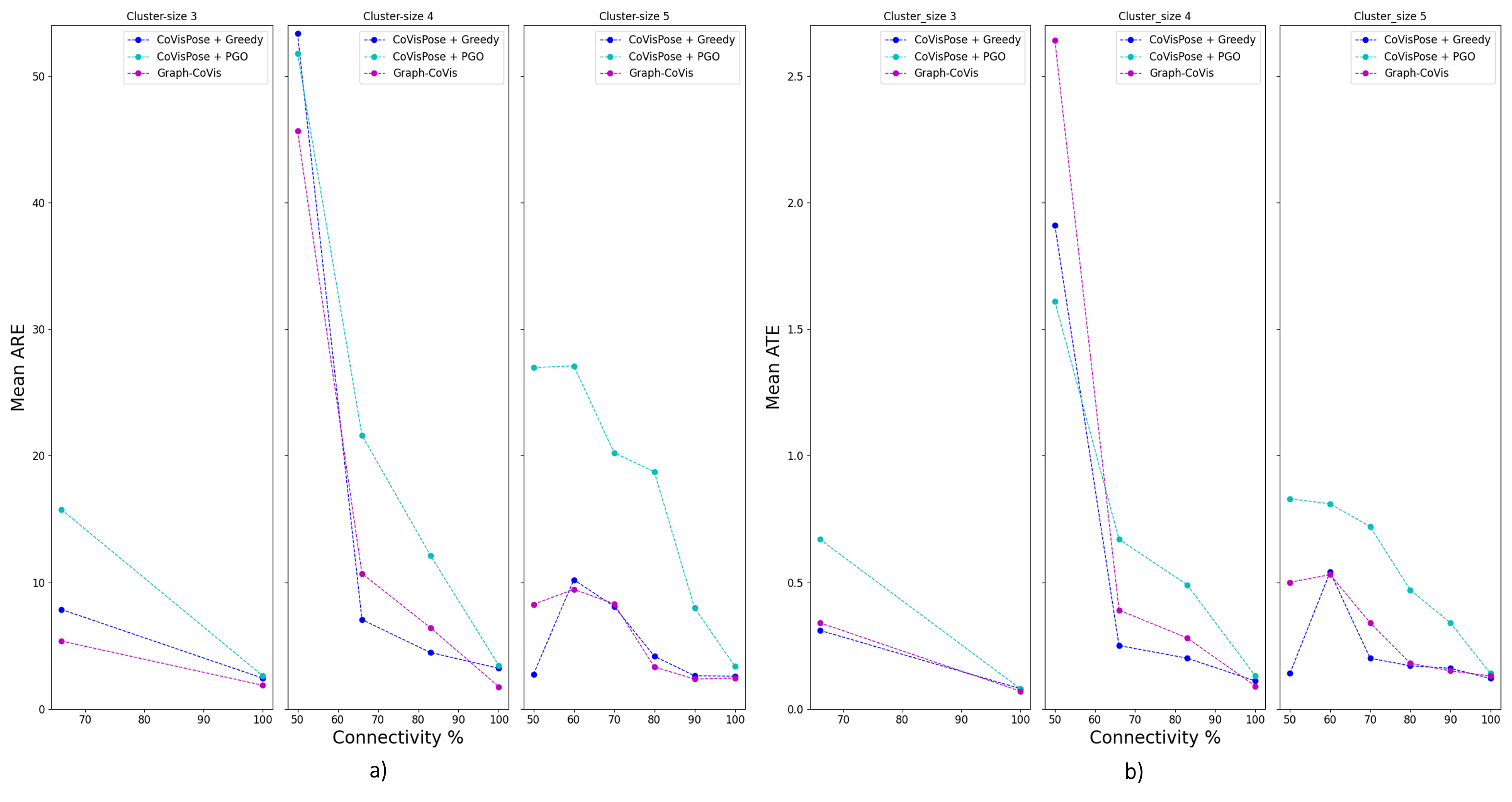}
    \caption{Connectivity percentage plotted against a) Mean ARE and b) Mean ATE in the graph for group size of three, four, and five.}
   \label{fig:connectivity-rotation-translation}
\end{figure*}

Figure \ref{fig:connectivity-rotation-translation} provides a visual representation of the ATE and ARE as a function of connectivity percentage for the three groups. As the connectivity percentage increases, we observe an improvement in the performance of Graph-CoVis as well as the baselines.

\begin{figure}[t]
  \centering
    \includegraphics[width=1\linewidth]{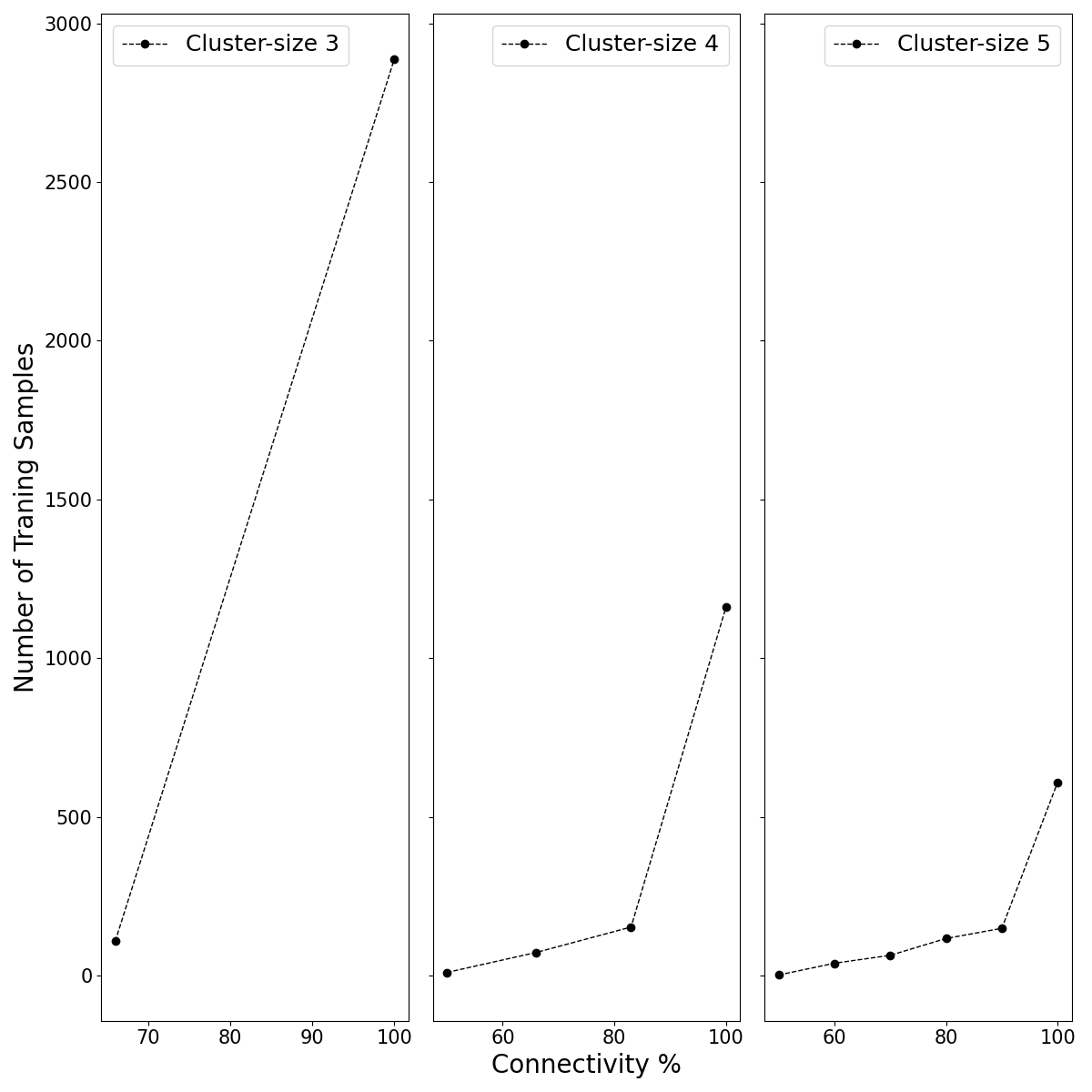}
    \caption{Connectivity percentage plotted against the number of training samples for the trained model, separated by group size of three, four, and five.}
   \label{fig:trainingvsconnectivity}
\end{figure}

Figure \ref{fig:trainingvsconnectivity} shows that the number of training examples is larger for higher connectivity percentages. While the baselines use an optimization step to obtain global poses, Graph-CoVis must learn the global poses in an end-to-end fashion, requiring sufficient training data across the spectrum of input cases. As such, we believe that increasing the number of training examples among the low connectivity percentage sets will benefit Graph-CoVis and further improve its performance.
\section{Qualitative examples}
\begin{figure*}
  \centering
    \includegraphics[width=0.87\linewidth]
    {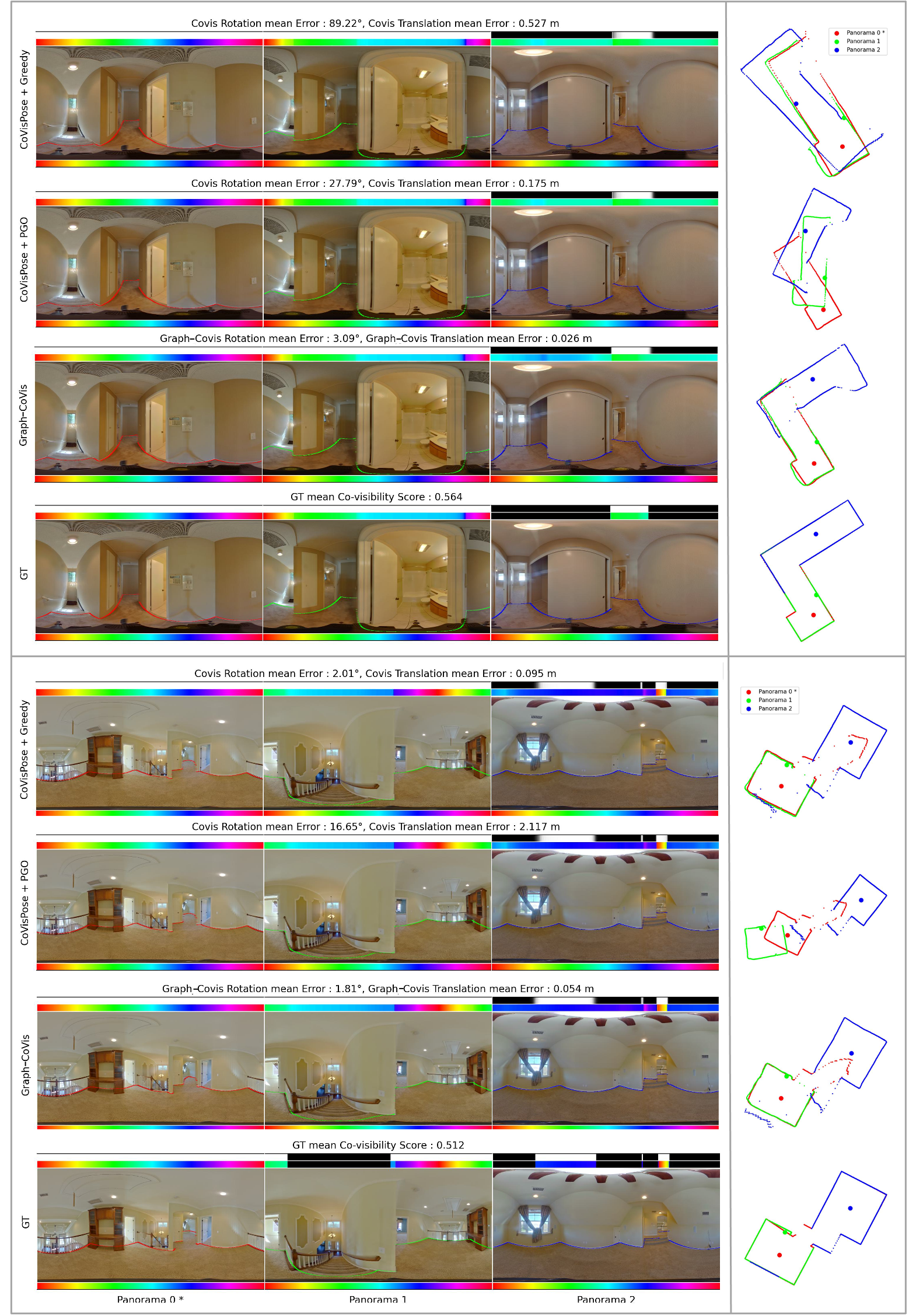}
    \caption{Two examples where Graph-CoVis performs significantly better than baselines.}
   \label{fig:Visualizations_significant}
\end{figure*}

\begin{figure*}
  \centering
    \includegraphics[width=0.87\linewidth]
    {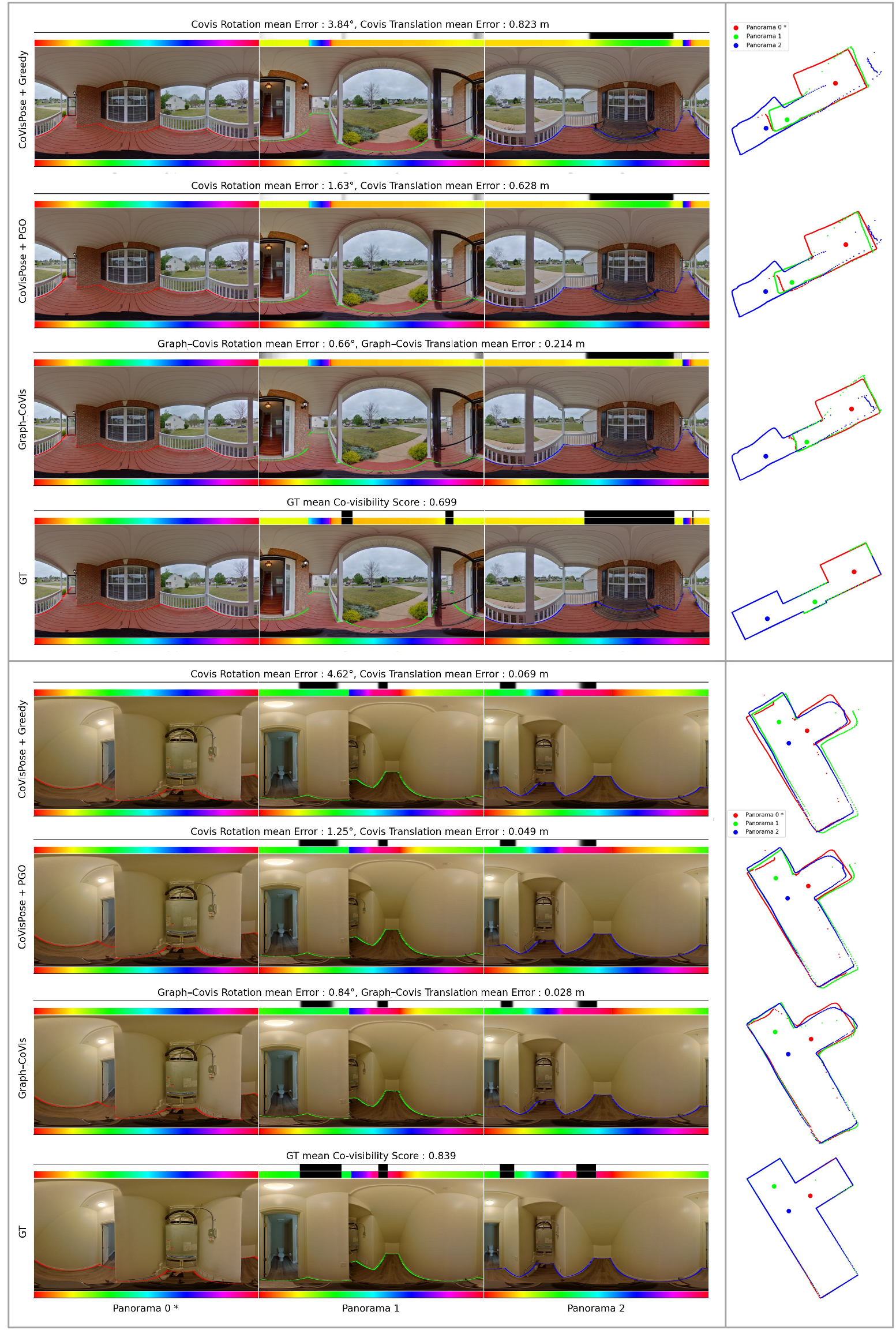}
    \caption{Two examples where Graph-CoVis performs moderately better than baselines }
   \label{fig:Visualizations_moderate}
\end{figure*}

\begin{figure*}
  \centering
    \includegraphics[width=0.87\linewidth]
    {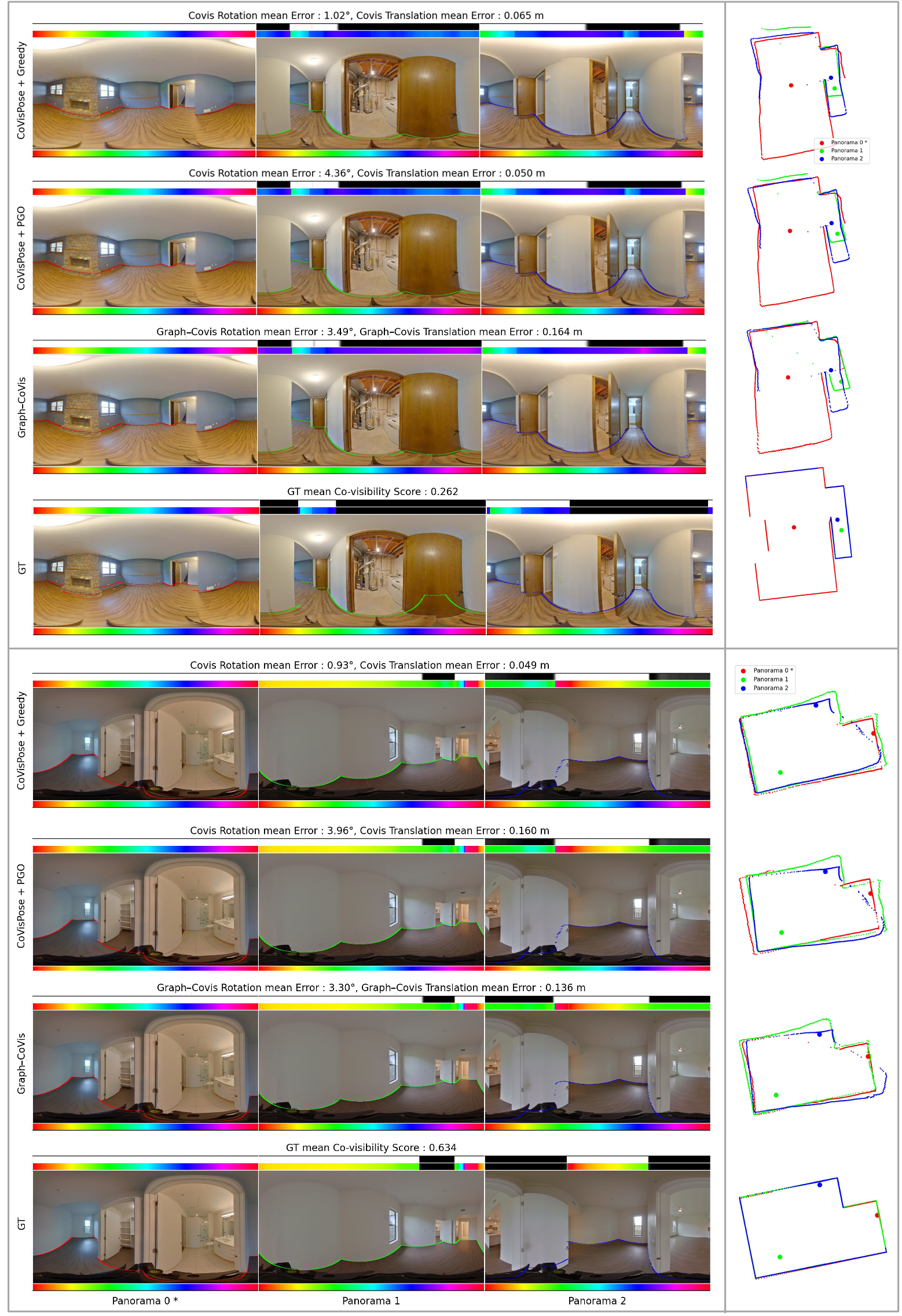}
    \caption{Two examples where Graph-CoVis performs worse than baselines. }
   \label{fig:covis-better}
\end{figure*}

Figures \ref{fig:Visualizations_significant} and \ref{fig:Visualizations_moderate} show some qualitative examples of where our system performs significantly better and moderately better than baseline approaches. 
Fig \ref{fig:covis-better} shows examples when our system performs worse than the baselines.

\section{Pose Graph Optimization}

We perform Pose Graph Optimization (PGO) using GTSAM\cite{Dellaert12_GTSAM}. We apply a diagonal Gaussian noise model on the prior constraint to specify the origin node, with standard deviations of $20$ cm and $0.1$ radians for translation and rotation, respectively. We also apply the same model for the odometry noise, with standard deviations of 30cm and .3 radians. We use the Levenberg-Marquardt optimizer with $1000$ iterations, with a relative error tolerance of $1\times10^{-5}$ for the convergence criteria. 







\section{Other Baselines}
A fair criticism may be made of our paper in that it could be possible to compare to a more extensive set of baselines based on additional recently published papers on multi-view image and panorama pose estimation. Notably, three recent works PoGO-Net\cite{Li2021PoGONetPG}, SALVe\cite{Lambert2022SALVeSA}, and Extreme SfM\cite{Shabani2021ExtremeSF} are relevant.


As explained in the paper, Extreme SfM and SALVe solve the problem of extreme-wide baseline pose estimation, subject to little-to-no visual overlap, to estimate floor level reconstruction of indoor spaces. Extreme SfM in particular focuses on the difficult cases where a single panorama is captured per room and ``\emph{seeks to align images from different rooms by exploiting the regularities of room arrangement at a house-scale}". Since our problem consists of multiple panoramas within the \emph{same large space and not at a house-scale}, we do not consider Extreme SfM as a directly comparable baseline approach.

SALVe is similar in application to Extreme SfM in its end goal of floor plan reconstruction, but by the nature of the method and input data distribution, it is a stronger candidate baseline. It handles all the panoramas captured in a floor of a home, which include multiple visually connected panoramas in a single space and visually not-connected panoramas across rooms. The former case is the same as our multi-view setting. SALVe uses separately trained depth and room layout estimation networks followed by a geometric alignment of the top-down projections of the room layouts. A deep network then verifies if the top-down projections are plausible. Finally a pose graph optimization algorithm estimates the global poses of all the panoramas that were connected by the alignment and verification steps. 
In theory, one could apply required modifications to the SALVe system to run on smaller sub-sets of panoramas once the SALVe code becomes available. 


Finally, PoGO-Net is a GNN-based alternative to pose graph optimization. Applied to perspective images, it requires a preprocessing step of
generating an initial view-graph that is subsequently refined and filtered by a GNN to estimate the final poses. 
A possible preprocessing approach is to estimate two-view poses using state-of-the-art for two-view panorama pose (which is CoVisPose) followed by pose graph optimization. This is indeed one of the chosen comparison baselines in our paper.
Upon PoGO-Net becoming publicly available, it's accuracy can be directly compared to Graph-Covis by applying it on the view-graph that results from running the CoVisPose + Greedy baseline.


{\small
\bibliographystyle{ieee_fullname}
\bibliography{egbib2}
}